%% file: Moreno_arxiv2016.tex
\documentclass[10pt,twocolumn,letterpaper]{article}

\input{newcommands2}
\usepackage{colortbl}
\usepackage{array}

\usepackage{cvpr}
\usepackage{times}
\usepackage{epsfig}
\usepackage{graphicx}
\usepackage{amsmath}
\usepackage{amssymb}
\usepackage{rotating}
\usepackage{upgreek}
\usepackage{balance}
\usepackage{bbding}

\usepackage{algorithm}
\usepackage{algorithmic}
\usepackage[utf8]{inputenc}

\usepackage{color}

\graphicspath{{Figures/}}

\usepackage[pagebackref=true,breaklinks=true,letterpaper=true,colorlinks,bookmarks=false]{hyperref}

\cvprfinalcopy 

\def\cvprPaperID{1034} 

\ifcvprfinal\fi
\begin{document}

\title{3D Human Pose Estimation from a Single Image via Distance Matrix Regression}

\author{Francesc Moreno-Noguer\\
Institut de Rob\`otica i Inform\`atica Industrial (CSIC-UPC), 08028, Barcelona, Spain\\
}

\maketitle

\begin{abstract}
This paper addresses the problem of 3D human pose estimation from a single image. We follow a standard two-step pipeline by   first detecting the 2D  position of the $N$ body joints, and then using these observations  to infer  3D pose.  For the first step, we use a recent CNN-based detector. For the second step, most existing approaches perform $2N$-to-$3N$ regression of the Cartesian joint coordinates. We show that more precise pose estimates can be obtained by   representing both the 2D and 3D human poses using $N\times N$ distance matrices, and formulating the problem as  a $2D$-to-$3D$ distance matrix regression.  For learning such a regressor we leverage on simple Neural  Network architectures, which by construction, enforce positivity and symmetry of the predicted  matrices. The approach has also the advantage to naturally handle  missing observations and allowing to hypothesize the position of  non-observed  joints. Quantitative results on  Humaneva and Human3.6M datasets demonstrate consistent performance gains over state-of-the-art. Qualitative evaluation on the images in-the-wild of the LSP dataset, using the regressor learned on Human3.6M, reveals very promising generalization results.  
\end{abstract}

\section{Introduction}

\begin{figure*}[t!]
\includegraphics[width=17.5cm, height=3.5cm]{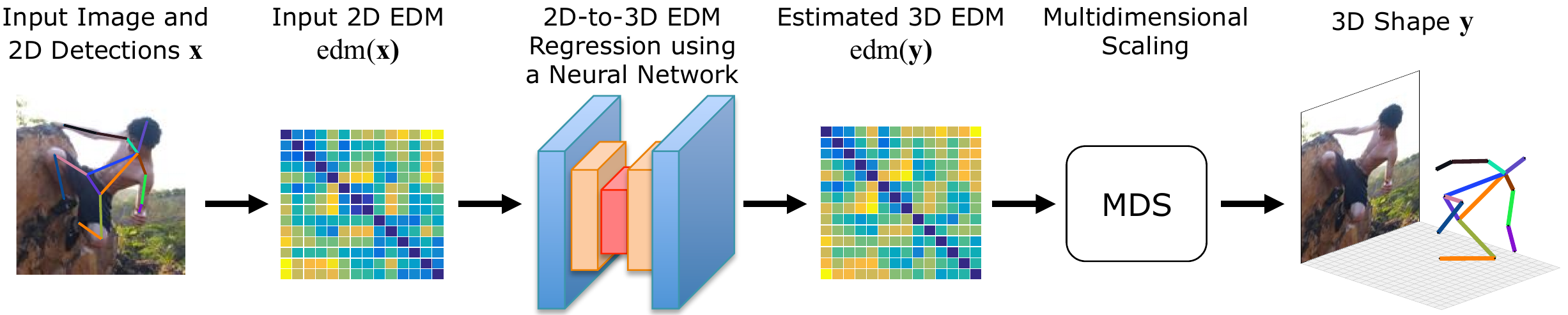}
\caption{{\bf Overview.} We formulate the 3D human pose estimation problem as a regression between two Euclidean Distance Matrices (EDMs), encoding pairwise distances of 2D and 3D body joints, respectively. The regression is carried out by a Neural Network, and the 3D joint estimates are obtained from the predicted 3D EDM via Multidimensional Scaling. } \label{fig:intro}
\end{figure*}

Estimating 3D human pose  from a single RGB image is known to be a severally ill-posed problem, because many different body configurations can virtually have  the same projection.  A typical solution consists in using discriminative strategies to directly learn mappings from image evidence (\eg HOG, SIFT) to 3D poses~\cite{AgarwalCVPR2004,RogezCVPR2008,SminchisescuCVPR2005,BoCVPR2008,ShakhnarovichCVPR2003}. This has been recently extended to end-to-end mappings using CNNs~\cite{LiICCV2015,TekinBMVC2016}.  In order to be effective, though, these approaches require large amounts of training images annotated with the ground truth 3D pose. While obtaining this kind of data is straightforward for  2D poses, even for images `in the wild' (\eg FLIC~\cite{SappCVPR2013} or LSP~\cite{JohnsonBMVC2010} datasets), it requires using sophisticated motion capture systems for the 3D case. Additionally, the datasets acquired this way (\eg Humaneva~\cite{SigalIJCV2010}, Human3.6M~\cite{IonescuPAMI2014}) are mostly indoors and their images are not representative of the type of image appearances outside the laboratory. 
 
 It seems therefore natural to split the problem in two stages: Initially use  robust image-driven 2D joint  detectors, and then  infer the  3D pose from these image observations using priors learned from mocap data. This pipeline   has already been used  in a number of works~\cite{SimoCVPR2012,SimoCVPR2013,YasinCVPR2016,WangCVPR2014,BogoECCV2016,RamakrishnaECCV2012} and is the strategy we consider in this paper. In particular, we first estimate 2D joints using a recent CNN detector~\cite{WeiCVPR2016}. For the second stage, however, most previous methods perform the 2D-to-3D inference in Cartesian space, between   $2N$- and  $3N$- vector representations of the  $N$ body joints. In contrast,  we propose representing  2D and 3D poses using $N\times N$ matrices of Euclidean distances between every pair of joints, and formulate the 3D pose estimation problem as one of 2D-to-3D distance matrix regression\footnote{Once the 3D distance matrix is predicted,  the position of the 3D body joints can be readily estimated using  Multidimensional Scaling (MDS).}.  Fig.~\ref{fig:intro}, illustrates our  pipeline. 
 
 Despite being extremely simple to compute, Euclidean Distance Matrices (EDMs) have several interesting advantages over vector representations that are particularly suited for  our problem. Concretely, EDMs: 1) naturally encode structural information of the   pose. Inference on vector representations needs to explicitly formulate such constraints; 2) are invariant to in-plane image rotations and translations, and  normalization operations  bring invariance to scaling;  3)  capture pairwise correlations and dependencies between all body joints.  

In order to learn a regression function that maps 2D-to-3D EDMs  we consider    Fully Connected (FConn) and Fully Convolutional (FConv) Network architectures. Since the dimension of our data is  small ($N\times N$ square matrices, with $N=14$ joints in our  model), input-to-output  mapping can be achieved through  shallow architectures, with only 2 hidden layers for the FConn and 4 convolutional layers for the FConv.  And most importantly, since the distance matrices used to train the networks are built from solely point configurations, we can easily synthesize artifacts and train the network under  2D detector noise and  body part occlusion. 




\comment{ 
In order to learn a regressor that maps 2D-to-3D EDMs  we use a  Fully Convolutional  Network (FCN).   Note that the size of the input and output matrices in our problem is much smaller than that for which FCNs are typically used~\cite{LongCVPR2015,FischerICCV2015}. We  will represent the human body with $N=14$ joints, yielding $14\times 14$ distance matrices. Mapping between such matrix spaces can be achieved through a shallow FCN architecture with  only two convolutional plus two deconvolutional layers. We will also introduce a last layer that enforces symmetry and positivity of the predicted matrices. Training will require relatively small number of samples and epochs, and testing (given the 2D detections) can be done in a fraction of a second on CPU. And most importantly, since the distance matrices used to train the network are built from solely point configurations, we can easily synthesize new data and train the network under novel viewpoints or simulate  artifacts such as 2D detector noise and  body part occlusion. 
}

We achieve state-of-the-art results on standard benchmarks including Humaneva-I and Human3.6M datasets,  and we show our approach to be robust to  large 2D detector errors, while (for the case of the FConv) also allowing to  hypothesize reasonably well occluded body limbs. Additionally, experiments in the Leeds Sports Pose dataset, using a network learned on Human3.6M, demonstrate good  generalization capability on images `in the wild'.

\section{Related Work}
Approaches to estimate 3D human pose from single images can be roughly split into two main categories:  methods that rely on generative models to constrain the space of possible shapes and discriminative approaches that directly predict 3D pose from image evidence. 

The most straightforward  generative model consists in representing human pose as linear combinations of modes learned from training data~\cite{BalanCVPR2007}. More sophisticated models allowing to represent larger deformations include spectral embedding~\cite{SminchisescuICML2004}, Gaussian Mixtures on Euclidean or Riemannian manifolds~\cite{HoweNIPS1999,SimoIJCV2016} and Gaussian processes~\cite{LawrenceICML2007,UrtasunCVPR2006,ZhaoTIP2011}. However, exploring the solution space defined by these generative models  requires iterative strategies and good enough initializations, making these methods more appropriate for tracking purposes.

Early discriminative approaches~\cite{AgarwalCVPR2004,RogezCVPR2008,SminchisescuCVPR2005,BoCVPR2008,ShakhnarovichCVPR2003} focused on directly predicting 3D pose from image descriptors such as SIFT of HOG filters, and more recently, from rich features encoding body part information~\cite{IonescuCVPR2014} and from the entire image in  Deep  architectures~\cite{LiICCV2015,TekinBMVC2016}. Since the mapping between feature  and pose space is complex to learn, the success of this family of techniques depends on the existence of large amounts of training images annotated with ground truth 3D poses. Humaneva~\cite{SigalIJCV2010} and Human3.6M~\cite{IonescuPAMI2014}, are two  popular MoCap  datasets used for this purpose. However, these datasets are acquired in  laboratory conditions, preventing the methods that uniquely use their data, to generalize well to unconstrained and realistic images. \cite{RogezNIPS2016} addressed this limitation by augmenting the training data for a CNN with automatically synthesized  images made of realistic textures.

Lying in between the two previous categories, there are a series of methods that first use discriminative formulations to estimate the 2D joint position, and then infer 3D pose using \eg  regression forests, Expectation Maximization or  evolutionary algorithms ~\cite{MoriPAMI2006,SimoCVPR2012,SimoCVPR2013,YasinCVPR2016,WangCVPR2014,BogoECCV2016,RamakrishnaECCV2012}. The two steps can be  iteratively  refined~\cite{SimoCVPR2013,YasinCVPR2016,WangCVPR2014} or formulated independently~\cite{RamakrishnaECCV2012,MoriPAMI2006,BogoECCV2016,SimoCVPR2012}. By doing this, it is then possible to exploit the full power current  CNN-based 2D detectors  like DeepCut~\cite{PishchulinCVPR2016} or the   Convolutional Pose Machines (CPMs)~\cite{WeiCVPR2016}, which have been trained with large scale datasets of images `in-the-wild'.



Regarding the 3D body pose parameterization, most  approaches use a skeleton with a number $N$ of joints  ranging between 14 and 20, and represented  by $3N$ vectors in a Cartesian space. Very recently, ~\cite{BogoECCV2016} used a generative volumetric model of the   full body. In order to enforce joint dependency during the 2D-to-3D  inference, ~\cite{IonescuPAMI2014} and ~\cite{TekinBMVC2016} considered latent joint representations, obtained  through Kernel Dependency Estimation and  autoencoders. In this paper, we propose using $N\times N$ Euclidean Distance Matrices for capturing such joint dependencies.

\begin{figure*}[t!]
\hspace{1mm}\includegraphics[width=17.0cm, height=3.7cm]{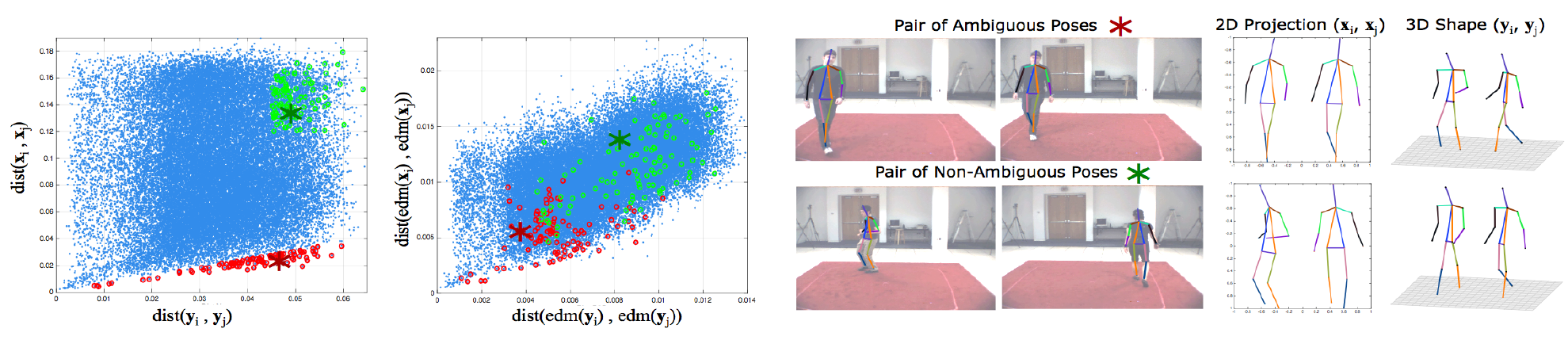}
\caption{{\bf EDMs vs Cartesian representations.} {\bf Left:} Distribution of relative 3D and 2D distances between random pairs of poses, represented as  Cartesian vectors (first plot) and EDM matrices (second plot). Cartesian representations show a more decorrelated pattern (Pearson correlation coefficient of $0.09$ against $0.60$ for the EDM), and in particular suffer from larger ambiguities, \ie poses with similar 2D projections and  dissimilar 3D shape.   Red circles indicate the most ambiguous  such poses, and green circles the most desirable configurations (large 2D and 3D differences). Note that red circles are more uniformly distributed along the vertical axis when using EDM representations, favoring larger differences and better discriminability. {\bf Right:}  Pairs of dissimilar 3D poses with similar (top)  and dissimilar (bottom) projections. They correspond to the dark red and dark green `asterisks' in the left-most plots. }
\label{fig:ambiguity}
\end{figure*}

EDMs have  already been used in similar domains, \eg in modal analysis  to estimate  shape basis~\cite{AgudoWACV2016},  to represent protein structures~\cite{KloczkowskiJSFG2009}, for sensor network localization~\cite{BiswasTASE2006} and for the resolution of kinematic constraints~\cite{PortaTR2005}. It is worth to point that for 3D shape recognition tasks, Geodesic Distance Matrices (GDMs) are preferred to EDMs, as they are  invariant to isometric deformations~\cite{SmeetsPR2012}. Yet, for the same reason, GDMs  are not suitable for our   problem, because multiple shape deformations  yield the same  GDM. In contrast,  the shape that produces a specific EDM is unique (up to  translation, rotation and reflection), and it can be estimated  via Multidimensional Scaling~\cite{BorgMDS2005,BiswasTASE2006}.

Finally, representing  2D and 3D joint positions by distance matrices, makes it possible to  perform inference with  simple Neural Networks. In contrast to recent CNN based methods  for 3D human pose estimation~\cite{LiICCV2015,TekinBMVC2016} we do not need to explicitly modify our networks to model the underlying joint dependencies.  This is directly encoded by the  distance matrices.

\section{Method}
Fig.~\ref{fig:intro} illustrates the main building blocks of our approach to estimate 3D human pose from a single RGB image. Given that image, we first detect body joints using a state-of-the-art detector. Then, 2D joints are normalized and represented by a  EDM, which is fed into a Neural Network to regress a EDM for the 3D body coordinates.  Finally, the position of the 3D joints is estimated via a `reflexion-aware' Multidimensional Scaling approach~\cite{BiswasTASE2006}.  We next describe in detail each of these steps.

\subsection{Problem Formulation}\label{sec:problem}
We represent the 3D  pose as a skeleton with $N\hspace{-1mm}=\hspace{-1mm}14$ joints and parameterized  by a $3N$ vector $\by=[\bp_1^\top,\ldots,\bp_N^\top]^\top$, where $\bp_i$ is the 3D location of the $i$-th joint. Similarly, 2D poses are represented by  2N vectors $\bx=[\bu_1^\top,\ldots,\bu_N^\top]^\top$, where $\bu_i $ are pixel coordinates. Given a full-body person image, our goal is to estimate the 3D  pose vector $\by$. For this purpose, we  follow a regression based discriminative approach. The most general formulation of this problem would involve using a set of training images  to learn a function that maps input images, or its features, to  3D poses. However, as discussed above, such a procedure would require a vast amount of data to obtain good generalization. 

Alternatively,  we will first compute the 2D joint position using the  Convolutional Pose Machine detector~\cite{WeiCVPR2016}. We denote by $\btx$  the output of the CPM, which is a noisy version of the ground truth 2D pose $\bx$. We also contemplate the possibility that some entries of $\btx$ are not observed due to joint occlusions or mis-detections. In order to not to change the dimension of $\btx$, the entries corresponding to these non-observed joints will be set to zero.

We can then formally write our problem as that of learning a mapping function $f:\mathbb{R}^{2N}\rightarrow \mathbb{R}^{3N}$ from  potentially corrupted 2D joint observations   $\btx$ to  3D poses $\by$,   given an annotated and clean training dataset  $\{\bx_i,\by_i\}_{i=1}^D$.


\begin{figure*}[t!]
\centering
\hspace{0mm}\includegraphics[width=15.3cm]{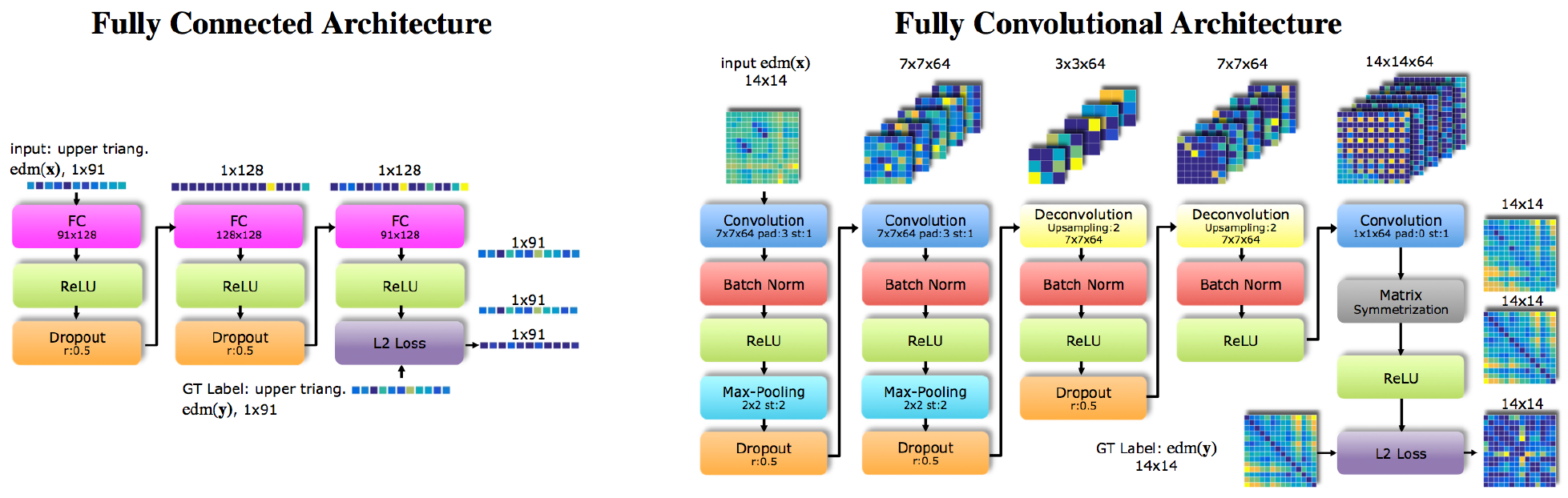}
\begin{tabular}{cc}
\end{tabular}
\caption{{\bf Neural Network Architectures used to perform 2D-to-3D regression of symmetric Euclidean Distance Matrices.}}
\label{fig:networks}
\end{figure*}

\subsection{Representing Human Pose with EDMs}


In oder to gain depth-scale invariance we first normalize the vertical coordinates of the projected 2D poses $\bx_i$ to be within the range $[-1,1]$. 3D joint positions $\by_i$ are expressed in meters with no further  pre-processing. We then represent both 2D and 3D poses by means of Euclidean Distance Matrices. For the 3D pose $\by$ we define $\textrm{edm}(\by)$ to be the $N\times N$ matrix where its $(m,n)$ entry is computed as:
\begin{equation}
\textrm{edm}(\by)_{m,n}=\|\bp_m-\bp_n \|_2\;.
\end{equation}
Similarly,  $\textrm{edm}(\bx)$ is  the $N\times N$ matrix built from the pairwise distances between  normalized 2D joint coordinates. 

Despite being simple to define, EDMs have several advantages over Cartesian  representations: EDMs are coordinate-free,  invariant to rotation, translation and reflection. Previous regression-based approaches~\cite{SimoCVPR2013,YasinCVPR2016,RogezNIPS2016}  need to compensate for this invariance  by pre-aligning the  training 3D poses   $\by_i$ w.r.t.  a global coordinate frame, usually defined by specific body joints.  Additionally, EDMs do not only  encode  the underlying structure of  plain 3D vector representations, but they also  capture richer information about pairwise correlations between all body joints. A direct consequence of both these advantages, is that EDM-based representations allow reducing the inherent ambiguities of the 2D-to-3D human pose estimation problem.


To empirically support this claim we randomly picked pairs of samples from the Humaneva-I dataset and plotted the distribution of  relative distances between their 3D and 2D  poses, using either Cartesian  or EDM representations (see Fig.~\ref{fig:ambiguity}).  For the Cartesian case (left-most plot), an entry to the graph corresponds to  $[\dist(\by_i,\by_j),\dist(\bx_i,\bx_j)]$, where $\dist(\cdot)$ is a normalized distance and $i, j$ are two random indices. Similarly, for the EDMs  (second plot), an entry to the graph corresponds to  $[\dist(\edm(\by_i),\edm(\by_j)),\dist(\edm(\bx_i),\edm(\bx_j))]$. Observe that 3D and 2D pairwise differences are much more correlated in this case. The interpretation of this pattern is that distance matrices yield larger 3D pose differences for most dissimilar 2D poses. The red circles in the graphs, correspond to the most ambiguous shapes, \ie, pairs of disimilar  poses $\{\by_i,\by_j\}$ with very  similar image projections $\{\bu_i,\bu_j\}$. Note that when using EDMs, these critical samples depict larger differences along the vertical axis, \ie, on the 2D representation.  This kind of behavior makes it easier the subsequent task of learning the 2D-to-3D mapping.

\begin{table*}[t!]
	\resizebox{17.4cm}{!} {
		\begin{tabular}
			{|l|cccc|cccc|cccc|} \hline
			\cellcolor[gray]{0.95}& \multicolumn{4}{|c|}{\cellcolor[gray]{0.95} \bf Walking (Action 1, Camera 1)}  & 	\multicolumn{4}{c|}{\cellcolor[gray]{0.95}\bf Jogging (Action 2, Camera 1)} &  \multicolumn{4}{c|}{\cellcolor[gray]{0.95}\bf Boxing (Action 5, Camera 1)} \\
{\cellcolor[gray]{0.95}\bf Method} & {\cellcolor[gray]{0.95}\bf S1} & {\cellcolor[gray]{0.95}\bf S2} & {\cellcolor[gray]{0.95}\bf S3} & {\cellcolor[gray]{0.95}\bf Average} &  {\cellcolor[gray]{0.95}\bf S1} & {\cellcolor[gray]{0.95}\bf S2} & {\cellcolor[gray]{0.95}\bf S3} & {\cellcolor[gray]{0.95}\bf Average} &  {\cellcolor[gray]{0.95}\bf S1} & {\cellcolor[gray]{0.95}\bf S2} & {\cellcolor[gray]{0.95}\bf S3} & {\cellcolor[gray]{0.95}\bf Average}\\
\hline\hline
Taylor CVPR'10~\cite{TaylorCVPR2010} &
48.80	& 47.40	& 49.80 &	48,70 &	75,35 &	- &	- &	- &	-& 	-& 	-& 	-\\ %
Bo  IJCV'10~\cite{BoIJCV2010} & %
45.40 &	28.30 & 62.30 &	45.33 & - &	- &	- &	- &	42.50 &	64.00 &	69.30&	58.60\\ %
Sigal  IJCV'12~\cite{SigalIJCV2012} & %
66.00 & 69.00 &	- &	- &	- &	- &	- &	- &	- &	- &	- &	-\\ %
Ramakrishna  ECCV'12~\cite{RamakrishnaECCV2012}(*) &
161.80 & 182.00 & 188.60 & 177.47 &	- &	- &	- &	- &	151.00 & 170.00 & 158.00 & 159.67\\ %
Simo-Serra  CVPR'12~\cite{SimoCVPR2012} & %
99.60 & 108.30 & 127.40 & 111.77 & - &	- &	- &	- &	- &	- &	- &	-\\ %
Simo-Serra  CVPR'13~\cite{SimoCVPR2013} & %
65.10 &  48.60 &  73.50 &  62.40 & 74.20 &  46.60 & 32.20 & 51.00 & - & - &  - &  -\\ %
Radwan  ICCV'13~\cite{RadwanICCV2013} & %
75.10 &	99.80 &	93.80 &	89.57 & 79.20 &	89.80 & 99.40 &	89.47 &	- &	- &	- &	- \\
Wang  CVPR'14~\cite{WangCVPR2014}& %
71.90 &	75.70 &	85.30 &	77.63 &	62.60 &	77.70 &	54.40 & 64.90 &	- &	- &	- &	-\\
Belagiannis  CVPR'14~\cite{BelagiannisCVPR2014}& %
68.30 &	- &	- &	- &	- &	- &	- &	- &	62.70 &	- &	- &	-\\
Kostrikov  BMVC'14~\cite{KostrikovBMVC2014}& %
44.00 &	30.90 &	41.70 &	38,87 &	57.20 &	35.00 &	33.30 &	41.83 &	- &	- &	- &	-\\
Elhayek  CVPR'15~\cite{ElhayekCVPR2015} & %
66.50 & - &	- &	- &	- &	- &	- &	- &	60.00 &	- & - &	-\\
Akhter  CVPR'15~\cite{AkhterCVPR2015}(*) & %
186.10 & 197.80 & 209.40 & 197.77 &	- &	- &	- &	- &	165.50 & 196.50 &	208.40 & 190.13 \\
Tekin  CVPR'16~\cite{TekinCVPR2016}& %
37.50 &	25.10 &	49.20 &	37.27 & - &	- &	- &	- &	50.50 &	61.70 &	57.50 &	56.57\\ %
Yasin  CVPR'16~\cite{YasinCVPR2016}  & %
35.80 &	32.40 &	41.60 &	36.60 & 46.60 & 41.40 &	35.40 &	41.13 &	- &	- &	- &	-\\ %
Zhou  PAMI'16~\cite{ZhouPAMI2016}(*) & %
100.00 & 98.90 & 123.10 & 107.33 &	- &	- &	- &	- &	112.50 & 118.60 & 110.00 &	113.70\\
Bogo  ECCV'16~\cite{BogoECCV2016} & %
73.30 &	59.00 & 99.40 &	77.23 &	- &	- &	- &	- &	82.10 &	79.20 &	87.20 &	82.83\\ %
\hline
\multicolumn{13}{|c|}{\cellcolor[gray]{0.95}\bf Our Approach, Fully Connected Network} \\
\hline
Train 2D: GT, Test: CPM & 35.70 & 36.80 & 41.34 & 37.95 & 41.25 & 27.96 &	34.34 &	34.52 &	47.26 &	50.52 &	67.64 & 55.14\\ %

Train 2D: CPM, Test: CPM & 20.16 & 14.00 &	28.76 &	20.97 &  38.12 &	17.95 &	21.42 &	25.83 &	44.05 &	48.52 &	57.00 &	49.86\\
Train 2D: GT+CPM, Test: CPM & %
19.72 &	13.52 &	26.46 &	19.90 &	{\bf 34.64} & {\bf 17.85} &	{\bf 20.05} &	{\bf 24.18} &	45.67 &	{\bf 47.52} &	57.63 &	50.27 \\ %
\hline
\multicolumn{13}{|c|}{\cellcolor[gray]{0.95}\bf Our Approach, Fully Convolutional Network} \\
\hline
Train 2D: GT, Test: CPM & %
28.35 &	27.75 &	38.93 &	31.68 &	47.75 &	27.82 &	30.21 &	35.26 &	{\bf 42.40} &	49.15 &	59.17 &	50.24\\  %
Train 2D: CPM, Test: CPM & %
19.82 &	{\bf 12.64} &	26.19 & 19.55 &	43.83 &	21.79 &	22.10 &	29.24 &	45.55 &	47.64 &	46.52 &	{\bf 46.57}\\
Train 2D: GT+CPM, Test: CPM & %
{\bf 19.68 } &	13.02 &	{\bf  24.89} & {\bf 19.20} &	39.69 &	20.04 &	21.04 & 26.92 & 46.63 &	47.56 &	{\bf 46.45} & 46.88\\
\hline
\end{tabular}}
\vspace{0.0mm}
\caption{{\bf Results on the Humaneva-I dataset.} Average error (in mm) between the ground truth and the predicted joint positions. `-' indicates that the results for that specific `action' and `subject' are not reported. The results of all  approaches are obtained from the original papers, except for (*), which were obtained from~\cite{BogoECCV2016}.}\label{tab:humaneva}
\end{table*}

\subsection{2D-to-3D Distance Matrix Regression}

The problem  formulated in Sec.~\ref{sec:problem} can now be rewritten in terms of finding the mapping $f:\mathbb{R}^{N\times N}\rightarrow \mathbb{R}^{N\times N}$, from potentially corrupted distance matrices $\edm(\btx)$ to matrices $\edm(\by)$ encoding the 3D pose, given a training set  $\{\edm(\bx_i),\edm(\by_i)\}_{i=1}^D$.   The expressiveness and  low dimensionality of the input and output data ($14\times14$ matrices) will make it possible to learn this mapping with relatively tiny Neural Network architectures, which we next describe. 

\vspace{1mm}
\noindent{\bf Fully Connected Network.} Since distance matrices are symmetric, we first consider a simple FConn architecture with 40K free parameters that regresses the $N(N-1)/2=91$ elements above the diagonal.   As shown in Fig.~\ref{fig:networks}-left, the network consists of three Fully Connected (FC) layers with 128-128-91 neurons. Each FC layer is followed by a rectified linear unit (ReLU). To reduce overfitting, we use
dropout after the first two layers, with a dropout ratio of 0.5 (50\% of probability to set a neuron's output value to zero). 

The $91$-dimensional vector at the output is used to build the $14\times 14$ output EDM, which by construction, is guaranteed to be symmetric. Additionally, the last ReLU layer enforces positiveness of all elements of the matrix, another necessary (but not sufficient) condition for an EDM.

\vspace{1mm}
\noindent{\bf Fully Convolutional Network.}
Motivated by the recent success of Fully Convolutional Networks in tasks like semantic segmentation~\cite{LongCVPR2015}, flow estimation~\cite{FischerICCV2015} and change dectection~\cite{AlcantarillaRSS2016}, we also consider the  architecture shown in Fig.~\ref{fig:networks}-right to  regress entire $14\times 14$ distance matrices. 

 FConv Networks were originally conceived to map  images or bi-dimensional arrays with some sort of spatial continuity. The EDMs however, do not convey this continuity and, in addition, they are defined up to a random permutation of the skeleton joints. In any event, for the case of human motion, the distance matrices turn to be  highly  structured,  particularly when handling  occluded joints, which results in patterns of zero columns and rows within the input EDM. In the experimental section we will show that FConv networks are also very effective for this kind of situations. 
 
Following previous works~\cite{AlcantarillaRSS2016,LongCVPR2015},  we explored an architecture with  a contractive and an expansive part. The contractive part consists of two convolutional blocks, with $7\times7$  kernels and $64$ features, each. Convolutional layers are followed by a Batch Normalization (BN) layer with learned  parameters, relieving from the task of having to compute such statistics from data during test. BN output is forwarded to a  non-linear ReLU; a $2\times 2$ max-pooling layer with stride 2 that performs the actual contraction; and finally, to a dropout layer with $0.5$ ratio.

The expansive part also has two main blocks which start with a deconvolution layer, that internally performs a $\times 2$ upsampling and, again, a convolution  with  $7\times7$  kernels and $64$ features. The deconvolution is followed by a ReLU and a dropout layer with ratio $0.5$. For the second block, dropout  is replaced by a convolutional layer that contracts the $64$, $14\times 14$ features into a single $14\times 14$ channel.

Note that there are no guarantees that the output of the expansive part will be a symmetric and positive matrix, as   expected for a EDM. Therefore, before computing the actual loss we designed a layer called  `Matrix Symmetrization' (MS) which enforces symmetry. If we denote by $\bZ$ the output of the expansive part,  MS will simply compute $(\bZ+\bZ^\top)/2$, which is symmetric. A final ReLU layer, guarantees that all values will be also positive.

This Fully Convolutional Network has 606K parameters.

\vspace{1mm}
\noindent{\bf Training.} In the experimental section we will report  results on multiple training setups. In all of them, the two networks were trained from scratch, and randomly initialized using the strategy proposed in~\cite{HeICCV2015}. We use a standard L2 loss function in the two cases. Optimization is carried out using Adam~\cite{KingmaICLR2015}, with a batch size of $7$ EDMs for Humaneva-I and $200$ EDMs for Human3.6M. FConn generally requires about 500 epochs to converge and FConv about 1500 epochs. We use default Adam parameters, except for the step size $\alpha$, which is initialized to $0.001$ and reduced to $0.0001$ after 250 (FConn) and 750 (FConv) epochs. The model definition and training is run under MatconvNet~\cite{VedaldiICM2015}.

\begin{figure*}[t!]
\begin{tabular}{ccccc}
{\bf \scriptsize Sample 3D Reconstructions} & %
{\bf \scriptsize Right Arm, Leg, True Tracks} & 
{\bf \scriptsize Right Arm, Leg, Hypoth. Tracks} & 
{\bf \scriptsize Left Arm, Leg, True Tracks} & 
{\bf \scriptsize Left Arm, Leg, Hypoth. Tracks} \\
\multicolumn{5}{c}{
\hspace{-0.0cm}\includegraphics[width=17.5cm, height=2.7cm]{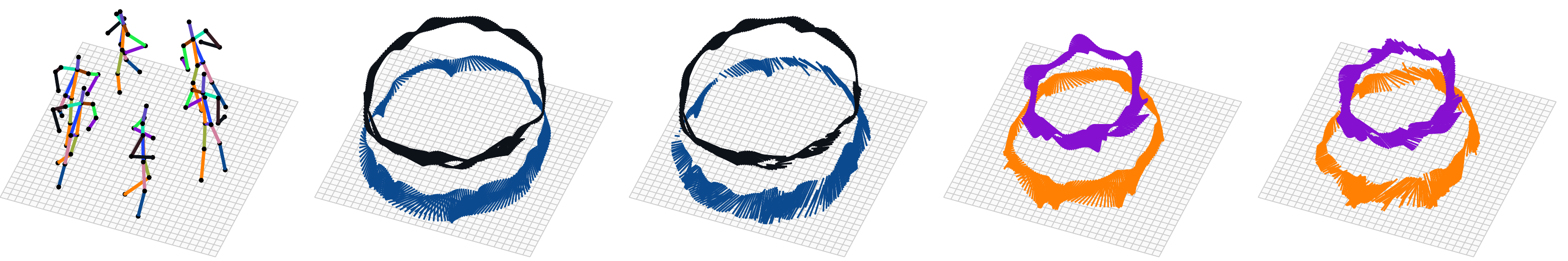}}
\end{tabular}
\caption{{\bf  Hypothesysing Occluded Body Parts.} Ground truth and hypothesized body parts obtained using the Fully Convolutional distance matrix regressor (Subject 3, action `Jogging'  from Humaneva-I). The network is trained with pairs of occluded joints and is able to predict one occluded limb (2 neighboring joints) at a time. Note how the generated tracks highly resemble the ground truth ones.
} \label{fig:occlusionhypotheses}
\end{figure*}

\begin{table*}[t!]
	\resizebox{17.4cm}{!} {
		\begin{tabular}
			{|lll|cccc|cccc|cccc|} \hline
			\multicolumn{3}{|l|}{\cellcolor[gray]{0.95}}& \multicolumn{4}{c|}{\cellcolor[gray]{0.95} \bf Walking (Action 1, Camera 1)}  & 	\multicolumn{4}{c|}{\cellcolor[gray]{0.95}\bf Jogging (Action 2, Camera 1)} &  \multicolumn{4}{c|}{\cellcolor[gray]{0.95}\bf Boxing (Action 5, Camera 1)} \\
			{\cellcolor[gray]{0.95}\bf NN Arch.} & {\cellcolor[gray]{0.95}\bf Occl. Type} & {\cellcolor[gray]{0.95}\bf Error}& {\cellcolor[gray]{0.95}\bf S1} &
			{\cellcolor[gray]{0.95}\bf S2} & {\cellcolor[gray]{0.95}\bf S3} & {\cellcolor[gray]{0.95}\bf Average} &  {\cellcolor[gray]{0.95}\bf S1} & {\cellcolor[gray]{0.95}\bf S2} & {\cellcolor[gray]{0.95}\bf S3} & {\cellcolor[gray]{0.95}\bf Average} &  {\cellcolor[gray]{0.95}\bf S1} & {\cellcolor[gray]{0.95}\bf S2} & {\cellcolor[gray]{0.95}\bf S3} & {\cellcolor[gray]{0.95}\bf Average}\\
			\hline\hline
			FConn & 2 Rand. Joints & Avg. Error & 53.30 & 59.49	& 51.99 &	54.93 &	49.93 &	27.91 &	37.07 &	38.30 &	49.63 &	60.74 &	64.53 &	58.30\\
			  &  &  Error Occl. Joints & 53.95 & 60.24 & 53.65 & 55.95 & 54.06 &	30.61 &	43.25 &	42.64 &	56.32 &	68.74 &	72.04 &	65.70\\
			FConn & Right Arm & Avg. Error & 55.55 & 59.16 &	49.31 &	54.67 &	54.45 &	31.30 &	36.73 &	40.83 &	49.48 &	69.59 &	68.26 &	62.44 \\
			 		&  &  Error Occl. Joints & 59.38 &	55.02 &	45.32 &	53.24 &	72.48 & 37.25 &	41.33 &	50.35 &	100.57 &	127.29 &	134.75 &	120.87 \\ 
		     FConn & Left Leg & Avg. Error & 53.51 &	55.87 &	60.04 &	56.47 &	49.83 &	30.90 &	41.16 &	40.63 &	46.68 &	64.85 &	62.04 &	57.86\\
					 &  &  Error Occl. Joints & 83.15 &	86.73 &	88.25 &	86.04 &	85.58 &	61.38 &	82.04 &	76.33 &	38.43 &	72.66 &	55.12 &	55.40\\ 
			\hline 		 
			FConv & 2 Rand. Joints & Avg. Error & 31.56& 	28.00 &	 38.49 &	32.68 &	46.63 &	26.61 &	34.34 & 35.86 &  50.00 & 54.19 & 56.12 &	53.44\\
			  &  &  Error Occl. Joints & 32.96 &	30.22 &	45.79 &	36.32 &	48.04 &	29.57 &	40.32 &	39.31 &	58.60 & 60.59 &	66.34 &	61.84\\
			FConv & Right Arm & Avg. Error & 37.96 & 27.70 &	35.27 &	33.64 &	51.00 &	28.11 &	31.81 &	36.97 &	58.09 &	59.74 &	62.78 &	60.20\\
			 			  &  &  Error Occl. Joints & 48.06 &	27.57 &	29.72 &	35.12 &	69.59 &	32.78 &	34.79 &	45.72 &	111.61 &	101.49 &	132.76	& 115.29\\
		     FConv & Left Leg & Avg. Error & 34.42 & 38.64 & 41.69 & 38.25 &	39.72 &	29.22 &	33.62 &	34.19 &	50.68 &	53.85 &	50.14 & 51.56\\
					 			  &  &  Error Occl. Joints & 61.03 &	64.79 &	71.48 &	65.77 &	59.3 &	 59.24 &	67.11 &	61.88 &	44.59 &	61.23 &	46.48 &	 50.77\\ 	\hline					 			  
\end{tabular}}
\vspace{0.0mm}
\caption{{\bf Results on the Humaneva-I under Occlusions.} Average overall joint error and average error of the occluded and hypothesized joints (in mm)   using the proposed  Fully Connected and  Fully Convolutional regressors. We train the two architectures with 2D GT+CPM and with random pairs of occluded joints. Test is carried out using the CPM detections with specific occlusion configurations.}\label{tab:humanevaocc}
\end{table*}

\subsection{From Distance Matrices  to 3D Pose}
Retrieving the 3D joint positions $\by=[\bp_1^\top,\ldots,\bp_N^\top]^\top$ from a potentially noisy distance matrix $\edm(\by)$ estimated by the neural network,   can be formulated as the following error minimization problem:
\begin{equation}
\argmin_{\bp_1,\ldots,\bp_N} \sum_{m,n} |\|\bp_m-\bp_n \|^2_2-\edm(\by)_{m,n} ^2|\;.  
\label{eq:mds} 
\end{equation}
We solve this minimization using~\cite{BiswasTASE2006},  a MDS algorithm  which poses  a semidefinite programming relaxation of the non-convex Eq.~\ref{eq:mds}, refined by a  gradient descent method.


Yet, note that the shape $\by$ we retrieve from $\edm(\by)$ is up to a reflection transformation, \ie,  $\by$ and its reflected version $\by^{*}$ yield the same distance matrix. In order to disambiguate this situation, we keep either $\by$ or $\by^{*}$ based on their degree of  anthropomorphism, measured as  the number of joints with angles  within the limits defined by the physically-motivated prior provided by~\cite{AkhterCVPR2015}. 


\section{Experiments}
We extensively evaluate the proposed approach on two publicly available datasets, namely Humaneva-I~\cite{SigalIJCV2010} and Human3.6M~\cite{IonescuPAMI2014}. Besides  quantitative comparisons w.r.t.  state-of-the-art  we also assess the robustness of our approach to noisy 2D observations and joint occlusion.  We furhter provide qualitative results on the LSP dataset~\cite{JohnsonBMVC2010} 

Unless specifically said, we assume the 2D joint positions in our approach are obtained with the CPM detector~\cite{WeiCVPR2016}, fed with a bounding box of the full-body person image.  As common practice in literature~\cite{SimoCVPR2013,YasinCVPR2016}, the reconstruction error we report refers to the average 3D Euclidean joint error (in mm), computed after rigidly aligning the estimated shape with the ground truth   (if available).

\subsection{Evaluation on Humaneva-I}

For the experiments with  Humaneva-I we  train our EDM  regressors on the  training sequences for the Subjects 1, 2 and 3, and evaluate on the `validation' sequences.  This is the same evaluation protocol used by the baselines we compare against~\cite{TaylorCVPR2010,BoIJCV2010,SigalIJCV2012,SimoCVPR2012,SimoCVPR2013,RamakrishnaECCV2012,RadwanICCV2013,WangCVPR2014,BelagiannisCVPR2014,KostrikovBMVC2014,ElhayekCVPR2015,AkhterCVPR2015,TekinCVPR2016,ZhouCVPR2016,ZhouPAMI2016,YasinCVPR2016,BogoECCV2016}. We report the performance on the `Walking', `Jogging' and `Boxing' sequences.  Regarding our own approach we consider several configurations depending on the type of regressor: Fully Connected or Fully Convolutional Network;  and depending on the type of 2D source used for training: Ground Truth (GT), CPM or GT+CPM. The 2D source used for evaluation is always CPM.

Table~\ref{tab:humaneva} summarizes the results, and shows that all configurations of our approach significantly outperform state-of-the-art. The improvement is particularly relevant when modeling the potential deviations of the CPM  by directly using  its  2D detections to  train the regressors. Interestingly, note that the results obtained by  FConn and FConv are very similar, being the former a much simpler architecture. However, as we will next show, FConv achieves remarkably  better  performance when dealing with occlusions.

\vspace{1mm}
\noindent{\bf Robustness to Occlusions.} The  3D pose we estimate obviously depends on the quality of the 2D CPM detections. Despite the CPM observations we have used do already contain certain errors, we have explicitly evaluated the robustness of  our approach under artificial  noise and occlusion artifacts. We next assess the impact of occlusions. We leave the study of the 2D noise for the following section.

\begin{table*}[t!]
	\resizebox{17.4cm}{!} {
		\begin{tabular}
			{|l|ccccccccccccccc|c|} \hline
{\cellcolor[gray]{0.95}\bf Method} & %
{\cellcolor[gray]{0.95}\bf Direct.} & %
{\cellcolor[gray]{0.95}\bf Discuss} & %
{\cellcolor[gray]{0.95}\bf Eat} & %
{\cellcolor[gray]{0.95}\bf Greet} & %
{\cellcolor[gray]{0.95}\bf Phone} & %
{\cellcolor[gray]{0.95}\bf Pose} & %
{\cellcolor[gray]{0.95}\bf Purch.} & %
{\cellcolor[gray]{0.95}\bf Sit} & %
{\cellcolor[gray]{0.95}\bf SitD} &  %
{\cellcolor[gray]{0.95}\bf Smoke} & %
{\cellcolor[gray]{0.95}\bf Photo} & %
{\cellcolor[gray]{0.95}\bf Wait} & %
{\cellcolor[gray]{0.95}\bf Walk}&
{\cellcolor[gray]{0.95}\bf WalkD} & %
{\cellcolor[gray]{0.95}\bf WalkT}&
{\cellcolor[gray]{0.95}\bf Avg} \\ %
\hline\hline 
\multicolumn{17}{|c|}{\cellcolor[gray]{0.98}\bf Protocol \#1}\\\hline
Ionescu PAMI'14~\cite{IonescuPAMI2014} & 132.71 & 183.55 & 133.37 & 164.39 & 162.12 & 205.94 &	150.61 & 171.31 & 151.57 &	243.03 & 162.14	& 170.69 & 177.13 & 96.60 & 127.88 & 162,20 \\
Li ICCV'15~\cite{LiICCV2015} & - &	136.88 & 96.94 & 124.74 &	-	 & 168.08 &	- &	- &	- &	- &	- &	- &	132.17 &	69.97 &	- & - \\
Tekin BMVC'16~\cite{TekinBMVC2016} & - & 129.06 &	91.43 &	121.68 &	- &	- &	- &	-& - &	-	&  162.17 &	- &	65.75 &	130.53 &	- & -\\ %
Tekin CVPR'16~\cite{TekinCVPR2016} & 102.41& 147.72 & 88.83 & 125.28 & 118.02 & 112.38 & 129.17 &	138.89 &	224.90 &	118.42 &	182.73 &	138.75 &	55.07 &	126.29 &	65.76 &	124.97\\
Zhou CVPR'16~\cite{ZhouCVPR2016} & 87.36 &	109.31 &	87.05 &	103.16	& 116.18 &	143.32 &	106.88 &	\bf{99.78} &	124.52 &	199.23 &	107.42	& 118.09 &	114.23	& \bf{79.39} &	97.70 & 112.91\\ %
Ours, FConv, Test 2D: CPM & \bf{69.54} &	\bf{80.15} &	\bf{78.20} &	\bf{87.01} &	\bf{100.75} &	\bf{76.01} &	\bf{69.65} &	104.71 & \bf{113.91} & \bf{89.68} & \bf{102.71} &	\bf{98.49} &	 \bf{79.18} &	82.40 &	\bf{77.17} & \bf{87.30}\\
\hline
\multicolumn{17}{|c|}{\cellcolor[gray]{0.98}\bf Protocol \#2}\\\hline
Ramakrishna ECCV'12~\cite{RamakrishnaECCV2012}(*) & 137.40  & 	149.30  & 	141.60  & 	154.30  & 	157.70  & 	141.80  & 	158.10  & 	168.60  & 	175.60  & 	160.40  & 	158.90  & 	161.70  & 	174.80  & 	150.00  & 	150.20  & 	156.03 \\
Akhter CVPR'15~\cite{AkhterCVPR2015}(*) & 1199.20  & 	177.60  & 	161.80  & 	197.80  & 	176.20	 &  195.40  & 	167.30  & 	160.70	  & 173.70  & 	177.80  & 	186.50 & 	181.90  & 	198.60 & 	176.20 & 	192.70 & 	181.56\\
Zhou PAMI'16~\cite{ZhouPAMI2016}(*) & 99.70 & 	95.80 & 	87.90 & 	116.80	 & 108.30 & 	93.50 & 	95.30 & 	109.10 & 	137.50	 & 106.00 & 	107.30 & 	102.20 & 	110.40 & 	106.50 & 	115.20	 & 106.10\\
Bogo ECCV'16~\cite{BogoECCV2016} & \bf{62.00} & 	\bf{60.20} & 	\bf{67.80} & 	\bf{76.50} & 	\bf{92.10}	 & \bf{73.00} & 	75.30	 & 100.30 & 	137.30 & 	\bf{83.40}& 	\bf{77.00} & 	\bf{77.30} & 	86.80 & 	\bf{79.70} & 	81.70 & 	\bf{82.03}\\
Ours. FConv, Test 2D: CPM & 66.07 & 	77.94 & 	72.58 & 	84.66 & 	99.71 & 	74.78 & 	\bf{65.29} & 	\bf{93.40} & 	\bf{103.14}	 & 85.03 & 	98.52 & 	98.78 & 	\bf{78.12} & 	80.05 & 	\bf{74.77}	 & 83.52\\
\hline
\multicolumn{17}{|c|}{\cellcolor[gray]{0.98}\bf Protocol \#3}\\\hline
Yasin CVPR'16~\cite{YasinCVPR2016} & 88.40 &	72.50&	108.50	& 110.20 &	97.10 &	81.60 &	107.20 &	119.00 &	170.80	& 108.20 &	142.50 &	86.90 &	92.10 &	165.70 &	102.00 &	110.18\\
Rogez NIPS'16~\cite{RogezNIPS2016} & -&	-&	-&	-&	-&	-&	-&	-&	-&	-&	-&	-&	-&	-&	-&	88.10\\
Ours, FConv, Test 2D: CPM & {\bf 67.44} &	{\bf 63.76} &	{\bf 87.15} &	{\bf 73.91} &	{\bf 71.48} &	{\bf 69.88} &	{\bf 65.08} &	{\bf 71.69} &	{\bf 98.63} &	{\bf 81.33} &	{\bf 93.25} &	{\bf 74.62} &	{\bf 76.51} &	{\bf 77.72} &	{\bf 74.63} &	{\bf 76.47} \\
\hline

\end{tabular}}
\vspace{0.0mm}
\caption{{\bf Results on the Human3.6M dataset.} Average joint error (in mm) considering the 3 evaluation prototocols described in the text.  The results of all  approaches are obtained from the original papers, except for (*), which are  from~\cite{BogoECCV2016}. 
 }\label{tab:human36m}
\end{table*}

\begin{table*}[t!]
	\resizebox{17.4cm}{!} {
		\begin{tabular}
			{|ll|ccccccccccccccc|c|} \hline
{\cellcolor[gray]{0.95}\bf Occ. Type} & %
{\cellcolor[gray]{0.95}\bf Error} &
{\cellcolor[gray]{0.95}\bf Direct.} & %
{\cellcolor[gray]{0.95}\bf Discuss} & %
{\cellcolor[gray]{0.95}\bf Eat} & %
{\cellcolor[gray]{0.95}\bf Greet} & %
{\cellcolor[gray]{0.95}\bf Phone} & %
{\cellcolor[gray]{0.95}\bf Pose} & %
{\cellcolor[gray]{0.95}\bf Purch.} & %
{\cellcolor[gray]{0.95}\bf Sit} & %
{\cellcolor[gray]{0.95}\bf SitD} &  %
{\cellcolor[gray]{0.95}\bf Smoke} & %
{\cellcolor[gray]{0.95}\bf Photo} & %
{\cellcolor[gray]{0.95}\bf Wait} & %
{\cellcolor[gray]{0.95}\bf Walk}&
{\cellcolor[gray]{0.95}\bf WalkD} & %
{\cellcolor[gray]{0.95}\bf WalkT}&
{\cellcolor[gray]{0.95}\bf Avg} \\ %
\hline\hline
2 Rnd.Joints & Avg. Error & 88.53	& 97.83 &	139.99 &	99.57 &	106.13 &	102.78 &	92.97 &	113.35 &	126.62 &	111.73 &	122.74 &	109.85 &	95.1 &	96.76 &	97.97	& 106.79\\
& Err.Occl. Joints & 94.77 &	104.37 &	155.66 &	110.48 &	119.62 &	103.83 &	91.04 &	141.31 &	135.35 &	137.76 &	146.68 &	131.41 &	116.16	 & 96.11 &	99.73 &	118.95 \\
Left Arm  & Avg. Error & 197.86 &	101.88 &	123.91 &	109.72 &	93.00 &	106.15 &	100.55	& 113.19	 & 129.50 &	111.15 &	135.72 &	118.07 &	99.21 &	100.73 &	100.94 &	109.44\\
& Err.Occl. Joints & 177.44 &	177.68 &	152.06 &	220.28 &	145.93 &	180.42 &	143.24 &	192.42 &	154.62 &	184.24 &	253.88 &	213.6 &	176.11 &	160.44 &	188.38 &	181.38 \\
Right Leg & Avg. Error & 79.94 &	82.23 &	132.64 &	92.05 &	100.77 &	97.32 &	76.37 &	126.95 &	125.51 &	106.66 &	109.82 &	95.92 &	94.88 &	89.82 &	91.60	&  100.17\\
& Err.Occl. Joints & 81.23&	92.57	& 177.80 &	103.69 &	148.45 &	120.74 &	92.63 &	200.56 &	183.03	& 146.10 &	145.29	& 107.36 &	133.11 &	105.9	& 120.12	& 130.57 \\\hline
\end{tabular}}
\vspace{0.0mm}
\caption{{\bf Results on  Human3.6M  under Occlusions.} Average overall joint error and average error of the hypothesized  occluded joints (in mm). The network is trained and evaluated according to the `Protocol \#3' described in the text. }\label{tab:human36mocc}
\end{table*}

\begin{table*}[t!]
	\resizebox{17.4cm}{!} {
		\begin{tabular}
			{|m{4.4cm}|ccccccccccccccc|c|} \hline
{\cellcolor[gray]{0.95}\bf 2D Input} & %
{\cellcolor[gray]{0.95}\bf Direct.} & %
{\cellcolor[gray]{0.95}\bf Discuss} & %
{\cellcolor[gray]{0.95}\bf Eat} & %
{\cellcolor[gray]{0.95}\bf Greet} & %
{\cellcolor[gray]{0.95}\bf Phone} & %
{\cellcolor[gray]{0.95}\bf Pose} & %
{\cellcolor[gray]{0.95}\bf Purch.} & %
{\cellcolor[gray]{0.95}\bf Sit} & %
{\cellcolor[gray]{0.95}\bf SitD} &  %
{\cellcolor[gray]{0.95}\bf Smoke} & %
{\cellcolor[gray]{0.95}\bf Photo} & %
{\cellcolor[gray]{0.95}\bf Wait} & %
{\cellcolor[gray]{0.95}\bf Walk}&
{\cellcolor[gray]{0.95}\bf WalkD} & %
{\cellcolor[gray]{0.95}\bf WalkT}&
{\cellcolor[gray]{0.95}\bf Avg} \\ %
\hline\hline
GT & 53.51&	50.52 &	65.76 &	62.47 &	56.9 &	60.63 &	50.83 &	55.95 &	79.62 &	63.68 &	80.83 &	61.80 &	59.42 &	68.53 &	62.11 &	62.17\\ 
GT+$\mathcal{N}(\bzero,5)$ & 57.05 &	56.05 &	70.33 &	65.46 &	60.39 &	64.49 &	59.06 &	58.62 &	82.80 &	67.85 &	83.97 &	70.13 &	66.76 &	75.04 &	68.62	& 67.11 \\ 
GT+$\mathcal{N}(\bzero,10)$ &  76.46 &	70.74 &	77.18 &	77.25 &	73.42 &	81.94 &	64.65 &	71.05 &	97.08 &	76.91 &	93.45 &	77.12 &	85.14 &	80.96 &	83.47	& 79.12\\
GT+$\mathcal{N}(\bzero,15)$ & 90.72	& 91.99 & 96.54 &	94.99 &	87.43 &	101.81 &	89.39 &	84.46 &	107.26 &	93.31 &	106.01 &	95.96 &	100.38 &	96.59 &	104.41 &	96.08\\ 
GT+$\mathcal{N}(\bzero,20)$ & 109.84 &	110.21 &	117.13 &	115.16 &	107.08 &	116.92 &	107.14 &	101.82 &	131.43 &	114.76 &	115.07 &	112.54	& 125.50 &	118.93 &	129.73 &	115.55 \\
\hline
\end{tabular}}
\vspace{0.0mm}
\caption{{\bf Results on the Human3.6M dataset under 2D Noise.} Average 3D joint error for increasing levels of 2D noise. The network is trained with 2D Ground Truth (GT) data and evaluated with GT+$\mathcal{N}(\bzero.\sigma)$. where $\sigma$ is the standard deviation (in pixels) of the noise.  }\label{tab:human36mnoise}
\end{table*}

\comment{
\begin{figure*}
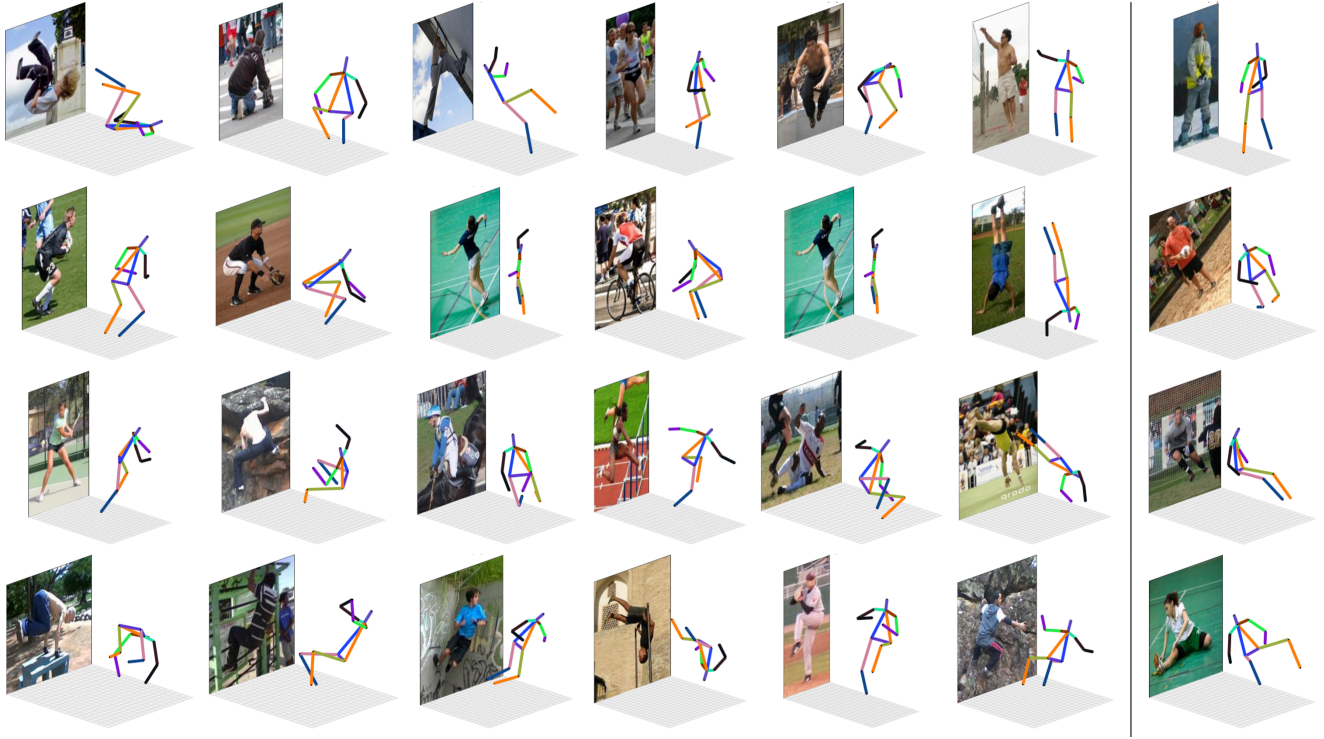

\centering
\begin{tabular}{cccccc|c}
\hspace{-2.5mm}\includegraphics[height=2.3cm]{Figures/Fig5/Good-cropped-CPM/im0007.pdf}&
\hspace{-2.5mm}\includegraphics[height=2.3cm]{Figures/Fig5/Good-cropped-CPM/im0011.pdf}&
\hspace{-2.5mm}\includegraphics[height=2.3cm]{Figures/Fig5/Good-cropped-CPM/im0012.pdf}&
\hspace{-2.5mm}\includegraphics[height=2.3cm]{Figures/Fig5/Good-cropped-CPM/im0017.pdf}&
\hspace{-2.5mm}\includegraphics[height=2.3cm]{Figures/Fig5/Good-cropped-CPM/im0358.pdf}&
\hspace{-2.5mm}\includegraphics[height=2.3cm]{Figures/Fig5/Good-cropped-CPM/im0102.pdf}&
\hspace{-0mm}\includegraphics[height=2.3cm]{Figures/Fig5/Fail-cropped-CPM/im0206.pdf} \\
\hspace{-2.5mm}\includegraphics[height=2.3cm]{Figures/Fig5/Good-cropped-CPM/im0078.pdf}&
\hspace{-2.5mm}\includegraphics[height=2.3cm]{Figures/Fig5/Good-cropped-CPM/im0083.pdf}&
\hspace{-2.5mm}\includegraphics[height=2.3cm]{Figures/Fig5/Good-cropped/im0148.pdf}&
\hspace{-2.5mm}\includegraphics[height=2.3cm]{Figures/Fig5/Good-cropped-CPM/im0153.pdf}&
\hspace{-2.5mm}\includegraphics[height=2.3cm]{Figures/Fig5/Good-cropped-CPM/im0148.pdf}&
\hspace{-2.5mm}\includegraphics[height=2.3cm]{Figures/Fig5/Good-cropped-CPM/im0167.pdf}&
\hspace{-0mm}\includegraphics[height=2.3cm]{Figures/Fig5/Fail-cropped-CPM/im0423.pdf}\\
\hspace{-2.5mm}\includegraphics[height=2.3cm]{Figures/Fig5/Good-cropped-CPM/im0170.pdf}&
\hspace{-2.5mm}\includegraphics[height=2.3cm]{Figures/Fig5/Good-cropped-CPM/im0221.pdf}&
\hspace{-2.5mm}\includegraphics[height=2.3cm]{Figures/Fig5/Good-cropped-CPM/im0251.pdf}&
\hspace{-2.5mm}\includegraphics[height=2.3cm]{Figures/Fig5/Good-cropped-CPM/im0851.pdf}&
\hspace{-2.5mm}\includegraphics[height=2.3cm]{Figures/Fig5/Good-cropped-CPM/im0475.pdf}&
\hspace{-2.5mm}\includegraphics[height=2.3cm]{Figures/Fig5/Good-cropped-CPM/im0568.pdf}& %
\hspace{-0mm}\includegraphics[height=2.3cm]{Figures/Fig5/Fail-cropped-CPM/im0560.pdf}\\
\hspace{-2.5mm}\includegraphics[height=2.3cm]{Figures/Fig5/Good-cropped-CPM/im0579.pdf}&
\hspace{-2.5mm}\includegraphics[height=2.3cm]{Figures/Fig5/Good-cropped-CPM/im0632.pdf}&
\hspace{-2.5mm}\includegraphics[height=2.3cm]{Figures/Fig5/Good-cropped-CPM/im1002.pdf}&
\hspace{-2.5mm}\includegraphics[height=2.3cm]{Figures/Fig5/Good-cropped-CPM/im1040.pdf}&
\hspace{-2.5mm}\includegraphics[height=2.3cm]{Figures/Fig5/Good-cropped-CPM/im1064.pdf}&
\hspace{-2.5mm}\includegraphics[height=2.3cm]{Figures/Fig5/Good-cropped-CPM/im1104.pdf}& %
\hspace{-0mm}\includegraphics[height=2.3cm]{Figures/Fig5/Fail-cropped-CPM/im0765.pdf}
\end{tabular}
\caption{{\bf Results on the LSP dataset.} The first six columns show correctly estimated poses. The right-most column shows failure cases. {}}\label{fig:lsp}
\end{figure*}
}

\begin{figure*}
\centering
\hspace{0mm}\includegraphics[width=17.5cm]{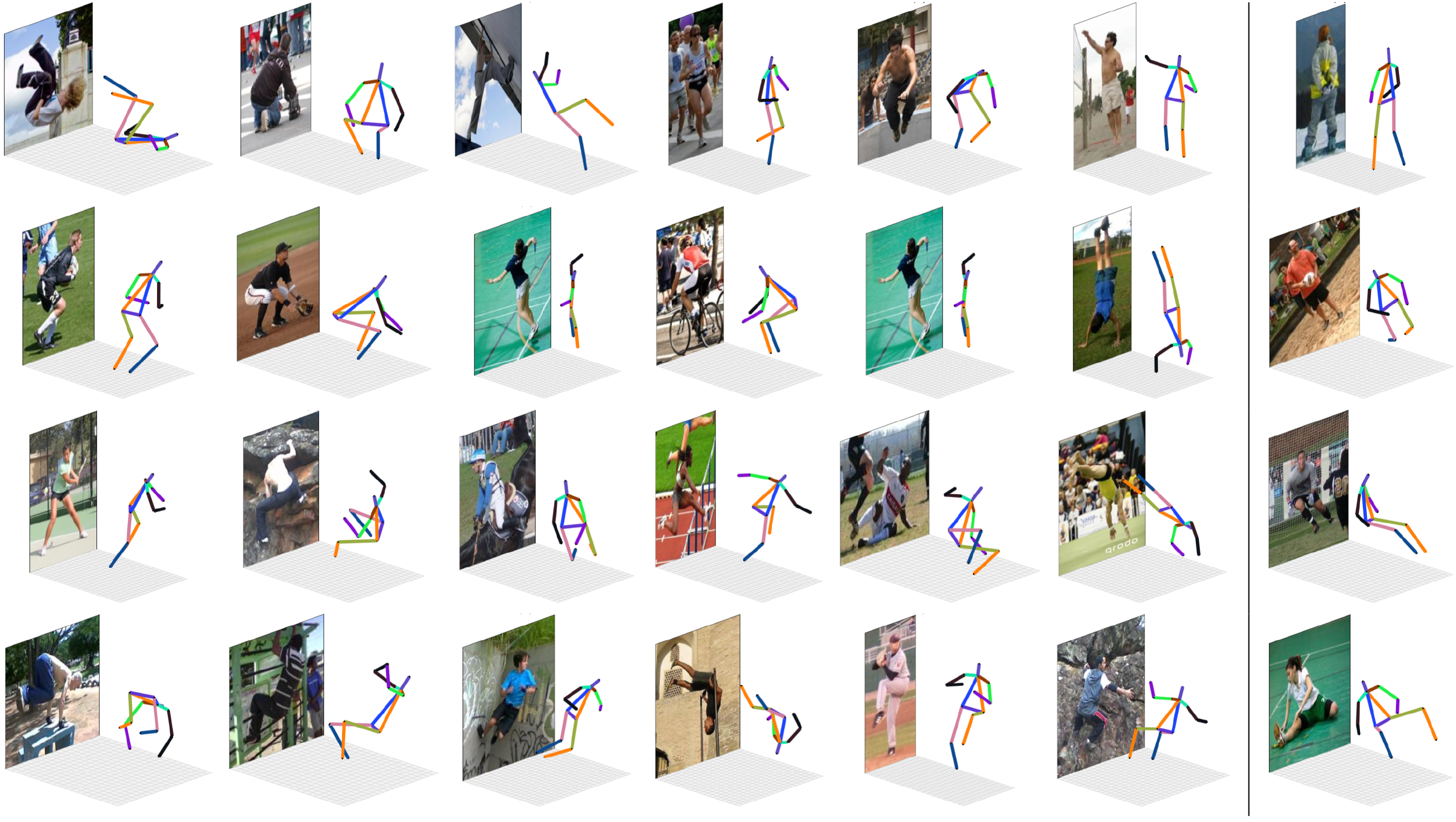}
\caption{{\bf Results on the LSP dataset.} The first six columns show correctly estimated poses. The right-most column shows failure cases. {}}\label{fig:lsp}
\end{figure*}

We consider two-joint occlusions, synthetically produced by removing two nodes of the projected skeleton. In order to make our networks robust to such occlusions and also able to hypothesize the 3D position of the non-observed joints, we re-train them using pairs $\{\edm(\bx_i^{\textrm{occ}}),\edm(\by_i)\}_{i=1}^D$, where $\bx_i^{\textrm{occ}}$ is the same as  $\bx_i$, but with  two random entries  set to zero. Note that this will make random rows and columns of $\edm(\bx_i^{\textrm{occ}})$ to be zero as well. At test, we consider  the cases with random joint occlusions  or with structured occlusions, in which we  completely remove the observation of one body limb (full leg  or full arm). 

Table~\ref{tab:humanevaocc} reports the reconstruction error for FConn and FConv.  Overall results show a clear and consistent advantage of the FConv network, yielding error values which are even comparable to state-of-the-art approaches when observing all joints (see Table~\ref{tab:humaneva}).  Furthermore, note that the error of the hypothesized joints is also within very reasonable bounds, exploiting only for the right arm position in the `boxing' activity. This is in the line of previous works which have shown that the combination of convolutional and deconvolutional layers  is very effective for image reconstruction and segmentation tasks~\cite{NohICCV2015,ZeilerICCV2011}. A specific example of joint hallucination is shown in Fig.~\ref{fig:occlusionhypotheses}.

\subsection{Evaluation on Human3.6M}

The Human3.6M dataset consists of 3.6 Million 3D poses of 11 subjects performing 15 different actions under 4  viewpoints. We found in the literature 3 different evaluation protocols. For Protocol \#1, 5 subjects (S1, S5, S6, S7 and S8) are used for training and 2 (S9 and S11) for testing. Training and testing is done independently per each action and using all camera views. Testing is carried out in all images. This  protocol  is  used in~\cite{IonescuPAMI2014,LiICCV2015,TekinCVPR2016,TekinBMVC2016,ZhouCVPR2016}. Protocol \#2 only differs from Protocol \#1 in that only the frontal view is considered for test. It has been recently used in~\cite{BogoECCV2016}, which also evaluates~\cite{RamakrishnaECCV2012,AkhterCVPR2015,ZhouPAMI2016}. Finally, in Protocol \#3, training data comprises all actions and viewpoints. Six subjects (S1, S5, S6, S7, S8 and S9) are used for training and every $64^{th}$ frame of the frontal view of S11 is used for testing. This is the protocol considered in~\cite{YasinCVPR2016,RogezNIPS2016}.

We will evaluate our approach on the  three protocols. However, since detecting the 2D  joints with CPM takes several seconds per frame, for  Protocols \#1 and \#2 we will test on every $8^{th}$  frame. For the same reason CPM detections will no longer be used during training, and   we will directly use  the  ground   truth 2D positions. For Protocol \#3 we  choose the training set by randomly picking 400K samples among all poses and camera views, a similar number as in~\cite{YasinCVPR2016}. In contrast to these works, no preprocessing is applied on the training set to maximize the pose variability. For the rest of experiments we will only consider the FConv regressor, which showed overall better performance than FConn  in the Humaneva dataset.

The results on Human3.6M are summarized in Table~\ref{tab:human36m}. For Protocols \#1 and \#3 our approach improves state-of-the-art by a considerable margin, and for Protocol \#2 is very similar to~\cite{BogoECCV2016}, a recent approach that relies on a high-quality volumetric prior of the body shape. 

\vspace{1mm}
\noindent{\bf Robustness to Occlusions.} We  perform the same occlusion analysis as we did for Humaneva-I and re-train the network under randomly occluded joints and test for random and structured  occlusions. The results (under Protocol \#3) are reported in Table~\ref{tab:human36mocc}. Again,  note that the average body error remains within  reasonable bounds. There are, however, some specific actions (\eg `Sit', `Photo') for which the occluded leg or arm are not very well hypothesized. We believe this is because in these actions, the limbs are in  configurations with only a few samples on the training set. Indeed, state of the art methods  also report poor performance on these actions, even when observing all joints. 
 
\vspace{1mm}
\noindent{\bf Robustness to 2D Noise.} We further analyze the robustness of our approach (trained on clean data) to  2D noise. For this purpose, instead of using CPM detections for test, we used the 2D ground truth test poses with increasing amounts of Gaussian noise. The results of this analysis are given in Table~\ref{tab:human36mnoise}. Note that the 3D  error gradually increases with the 2D noise, but does not seem to break the system. Noise levels of up to 20 pixels std are still reasonably supported. As a reference, the mean 2D error of the CPM detections considered in Tables~\ref{tab:human36mocc} and~\ref{tab:human36m} is of $10.91$ pixels. Note also that there is still room for improvement, as  more precise 2D detections   can considerably boost the  3D pose accuracy.

\subsection{Evaluation on Leeds Sports Pose Dataset}
We finally explore the generalization capabilities of our approach  on the  LSP dataset. For each  input image, we  locate  the 2D joints using the CPM detector,  perform the 2D-to-3D EDM regression using the Fully Convolutional network learned on Human3.6M (Protocol \#3) and compute the 3D pose through MDS. Additionally, once the 3D pose is estimated, we retrieve the rigid rotation and translation that aligns it with the input image using a PnP algorithm~\cite{LuPAMI2000}. Since the internal parameters of the camera are unknown, we sweep the focal length configuration space and keep the solution that minimizes the reprojection error.  

The lack of 3D annotation makes it not possible to perform a quantitative evaluation of the 3D shapes accuracy. Instead, in Table~\ref{tab:lsp}, we report three types of 2D reprojection errors per body part, averaged over the 2000 images of the dataset: 1) Error of the CPM detections; 2) Error of the reprojected shapes when estimated using  CPM 2D detections;  and 3) Error of the reprojected shapes when estimated using 2D GT annotations. While these results do not guarantee good accuracy of the estimated shapes, they are indicative that the method is working properly.  A visual inspection of the  3D estimated poses, reveals very promising results, even for poses which do not appear on the Human3.6M dataset used for training (see Fig.~\ref{fig:lsp}).  There still remain  failure cases (shown on the right-most column), due to \eg detector mis-detections, extremal body poses or camera viewpoints that largely differ from those of  Human3.6M. 


\begin{table}[t!]
	\resizebox{8.3cm}{!} {
		\begin{tabular}
			{|l|cccccccc|c|} \hline
{\cellcolor[gray]{0.95}\bf Error Type} & %
{\cellcolor[gray]{0.95}\bf Feet} & %
{\cellcolor[gray]{0.95}\bf Knees} & %
{\cellcolor[gray]{0.95}\bf Hips} & %
{\cellcolor[gray]{0.95}\bf Hands} & %
{\cellcolor[gray]{0.95}\bf Elbows} & %
{\cellcolor[gray]{0.95}\bf Should.} & %
{\cellcolor[gray]{0.95}\bf Head} & %
{\cellcolor[gray]{0.95}\bf Neck} & %
{\cellcolor[gray]{0.95}\bf Avg} \\ %
\hline\hline
CPM & 5.66&	4.22 &	4.27 &	7.25 &	5.24 &	3.17 &	3.55 &	2.65 &	4.77\\
Reproj. 2D CPM & 12.9 &	9.21 &	9.52 &	10.72  &	8.27 &	5.8 &	8.18 &	5.47&	9.08\\
Reproj. 2D GT & 9.78 &	6.88 &	7.83 &	6.48 &	6.38 &	4.53 &	6.34 &	4.14	& 6.76\\
\hline
\end{tabular}}
\vspace{0.0mm}
\caption{{\bf Reprojection Error (in pixels) on the LSP dataset.}  }\label{tab:lsp}
\vspace{-2mm}
\end{table}

\section{Conclusion}
In this paper we have formulated the 3D human pose estimation problem as a regression between matrices encoding 2D and 3D joint distances. We have shown that such matrices help to reduce the inherent ambiguity of the problem by naturally incorporating structural information of the human body and capturing  joints correlations. The distance matrix regression is carried out by simple Neural Network architectures, that bring robustness to  noise and occlusions of the 2D detected joints. In the latter case, a Fully Convolutional network has allowed to hypothesize unobserved body parts. Quantitative evaluation on standard benchmarks shows remarkable improvement compared to state of the art. Additionally, qualitative results on images `in the wild' show the approach to generalize well to untrained data. Since distance matrices just depend on   joint positions, new training data from novel viewpoints and shape configurations can be readily synthesized. In the future, we plan to explore online training strategies  exploiting this.

\section{Acknowledgments}
This work is partly funded by the Spanish MINECO project RobInstruct TIN2014-58178-R and by the ERA-Net Chistera project I-DRESS PCIN-2015-147. The author thanks Nvidia for the hardware donation under the GPU grant program, and   Germán Ros for fruitful discussions that initiated this work.

\comment{ 
***Gall CVPR 2016***: 
Discriminative methods:
[1]Agarwal CVPR 2004. 
[2]Agarwal PAMI 2006
[6] Bo IJCV 2008
[7] Bo CVPR 2008
[18] Mori PAMI 2006
[27] Sminchisescu CVPR 2005

CNN: 
[16] Li ACCV 2015
[17] Li ICCV 2015

Generative + Discriminative
[14]  3D PSM. Lostrikov

Mocap:
[21] Ramakrishna ECCV 2012
[25] Simo CVPR 2013
[26] Simo CVPR 2012
[29] Wang CVPR 2014
[33] Yasin MIRAGE 2013

*** LEpetit***
generative models for plausible configuration align with image evidence:
[12] Gall IKCV 2010
[13] Gammeter ECCV 2008
[27] Ormoneit WHMAS 2000
[35] Sidenbladh 2000

3D pictorial
[3] Belagiannis CVPR 2014
[6] Bruenius CVPR 2013

Discriminative regression based: direct mapping from image evidence to 3D poses
[1] Agarwal CVPR 2004
[4] Bo IJCV 2010
[16] Ionescu CVPR 2014
[40] Sminchisescu CVPR 2005

****Sminchisescu CVPR 2014***
Discriminative
[1] Agarwal CVPR 2004
[17] Ionescu ICCV 2011
[20] Kanaujia CVPR2007
[28] Sigal NIPS 2007
[31] Smincheisescu CVPR 2005

}

\comment{
FLIC: B. Sapp and B. Taskar. Modec: Multimodal decomposable models for human pose estimation. In CVPR. 2013.

LSD: S. Johnson and M. Everingham. Clustered pose and nonlinear appearance models for human pose estimation. In BMVC. 2010.

MPII: M. Andriluka. L. Pishchulin. P. Gehler. and B. Schiele. 2D human pose estimation: New benchmark and state of the art analysis. In CVPR. 2014.

A. Balan. L. Sigal. M. Black. J. Davis. H. Haussecker. De-
tailed Human Shape and Pose from Images. In CVPR. 2007.

C. Sminchisescu. A. Kanaujia. Z. Li. and D. Metaxas. Dis- criminative Density Propagation for 3D Human Motion Es- timation. In CVPR. 2005.

G. Mori and J. Malik. Recovering 3d human body configurations using shape contexts. IEEE Transactions on Pattern Analysis and Machine Intelligence. 28(7). 2006.

G. Shakhnarovich. P. Viola. and T. Darrell. Fast pose estimation with parameter- sensitive hashing. In IEEE Conference on Computer Vision and Pattern Recognition. 2003.

Ramakrishna. V.. Kanade. T.. Sheikh. Y.: Reconstructing 3D human pose from 2D image landmarks. In: European Conference on Computer Vision. ECCV. pp. 573–586 (2012)

Loper. M.. Mahmood. N.. Romero. J.. Pons-Moll. G.. Black. M.J.: SMPL: A
skinned multi-person linear model. ACM Transactions on Graphics (TOG) - Pro-
ceedings of ACM SIGGRAPH Asia 34(6). 248:1–248:16 (2015)

I. Borg and P. Groenen. Modern Multidimensional Scaling: Theory and Applications. Springer. 2005.

[1] P. Biswas. T.-C. Liang. K.-C. Toh. T.-C. Wang. and Y. Ye.    
  Semidefinite programming approaches for sensor network localization 
  with noisy distance measurements. 
  IEEE Transactions on Automation Science and Engineering. 
 3 (2006). pp. 360--371.

}

\balance

{\small
\bibliographystyle{ieee}
\bibliography{string,references}
}

\end{document}

%% file: newcommands2.tex
\newcommand{\comment}[1]{}

\newcommand{\edm}{\textrm{edm}}
\newcommand{\dist}{\textrm{dist}}

\newcommand{\bzero}{\textbf{0}}

\newcommand{\bp}{\mathbf{p}}

\newcommand{\bu}{\mathbf{u}}

\newcommand{\bx}{\mathbf{x}}
\newcommand{\btx}{\tilde{\mathbf{x}}}
\newcommand{\by}{\mathbf{y}}

\newcommand{\bZ}{\mathbf{Z}}

\newcommand{\argmin}{\operatornamewithlimits{arg\,min}}